\documentclass[conference]{IEEEtran}

\ifCLASSINFOpdf
  \usepackage[pdftex]{graphicx}
  \graphicspath{/figures}
  \DeclareGraphicsExtensions{.pdf,.jpeg,.png}
\else
\fi

\usepackage{amsmath}
\usepackage{url}

\begin{document}
\title{iFAN Ecosystem: A Unified AI, Digital Twin, Cyber‑Physical Security, and Robotics Environment for Advanced Nuclear Simulation and Operations}

\author{
\IEEEauthorblockN{Youndo Do, Chad Meece, Marc Zebrowitz, Spencer Banks, \\
Myeongjun Choi, Xiaoxu Diao, Kai Tan, Michael Doran, Jason Reed, Fan Zhang}
\IEEEauthorblockA{
Georgia Institute of Technology\\
Atlanta, Georgia 30332, USA
}
}

\maketitle

\begin{abstract}
As nuclear facilities experience digital transformation and advanced reactor development, AI integration, cyber-physical security, and other emerging technologies such as autonomous robot operations  are increasingly developed. However, evaluation and deployment is challenged by the lack of dedicated virtual testbeds. The Immersive Framework for Advanced Nuclear (iFAN) ecosystem is developed, a comprehensive digital twin framework with a realistic 3D environment with physics-based simulations. The iFAN ecosystem serves as a high-fidelity virtual testbed for plant operation, cybersecurity, physical security, and robotic operation, as it provides real-time data exchange for pre-deployment verification. Core features include virtual reality, reinforcement learning, radiation simulation, and cyber-physical security. In addition, the paper investigates various applications through potential operational scenarios. The iFAN ecosystem provides a versatile and secure architecture for validating the next generation of autonomous and cyber-resilient nuclear operations.
\end{abstract}


%
\IEEEpeerreviewmaketitle

\section{Introduction}

The global transition of modernizing, new construction, re-purposing legacy nuclear facilities along with the emergence of advanced reactors has necessitated a paradigm shift in nuclear operations. This evolution is driven by the escalating energy requirements of large-scale artificial intelligence (AI) infrastructure. As advanced reactor designs approach the deployment stages, there is a demand to move from static, fixed-sensor monitoring toward autonomous robotic platforms or sensor networks. These systems are essential for navigating the high-consequence, high-radiation environments where human intervention is limited.

However, the deployment of such systems presents two critical concerns. First, as these robotic platforms are not yet fully operational in standardized nuclear workflows, there is an urgent need for high-fidelity virtual testbeds. Digital twins (DT) mitigate this issue and provide a safe and cheaper alternative to validate developments in robotic path planning, interaction within the nuclear power plant (NPP) environment, and sensor integration before physical prototypes and testing. Second, as these robots and installed sensors are increasingly networked, they introduce an expanded attack surface.

To address these concerns, a framework must be developed for robot platforms and monitoring equipment in NPPs that is both robust and secure. This paper proposes a unified architecture that evolves into a comprehensive DT Ecosystem. While iFANnpp \cite{do2026ifannpp}, the predecessor of this work, focused primarily on the control validation and operational feasibility of robotic platforms, this project expands that scope by integrating more functionality as well as a cybersecurity layer. The proposed ecosystem serves as a dual-purpose solution: acting as a pre-deployment verification for evolving robotic platforms while simultaneously employing a cyberspace testbed to detect, defend against, and mitigate cyber-attacks.

Distinct from conventional DT implementations that primarily synchronize three-dimensional (3D) geometric models with real-world assets, the architecture proposed in this project integrates those models with a high-fidelity, full-scope, physics-based NPP simulator, called Generic Pressurized Water Reactor (GPWR) \cite{gpwr}. Similar to traditional DTs, the DT ecosystem has a real-time data exchange with the simulator, which justifies the ecosystem for use in the pre-deployment phase. This integration ensures that the ecosystem provides an authoritative virtual testbed, allowing for the rigorous testing of both robot operations and cyber-resilience in the pre-deployment phase.

\begin{figure*}[t]
    \centering
    \includegraphics[width=\linewidth]{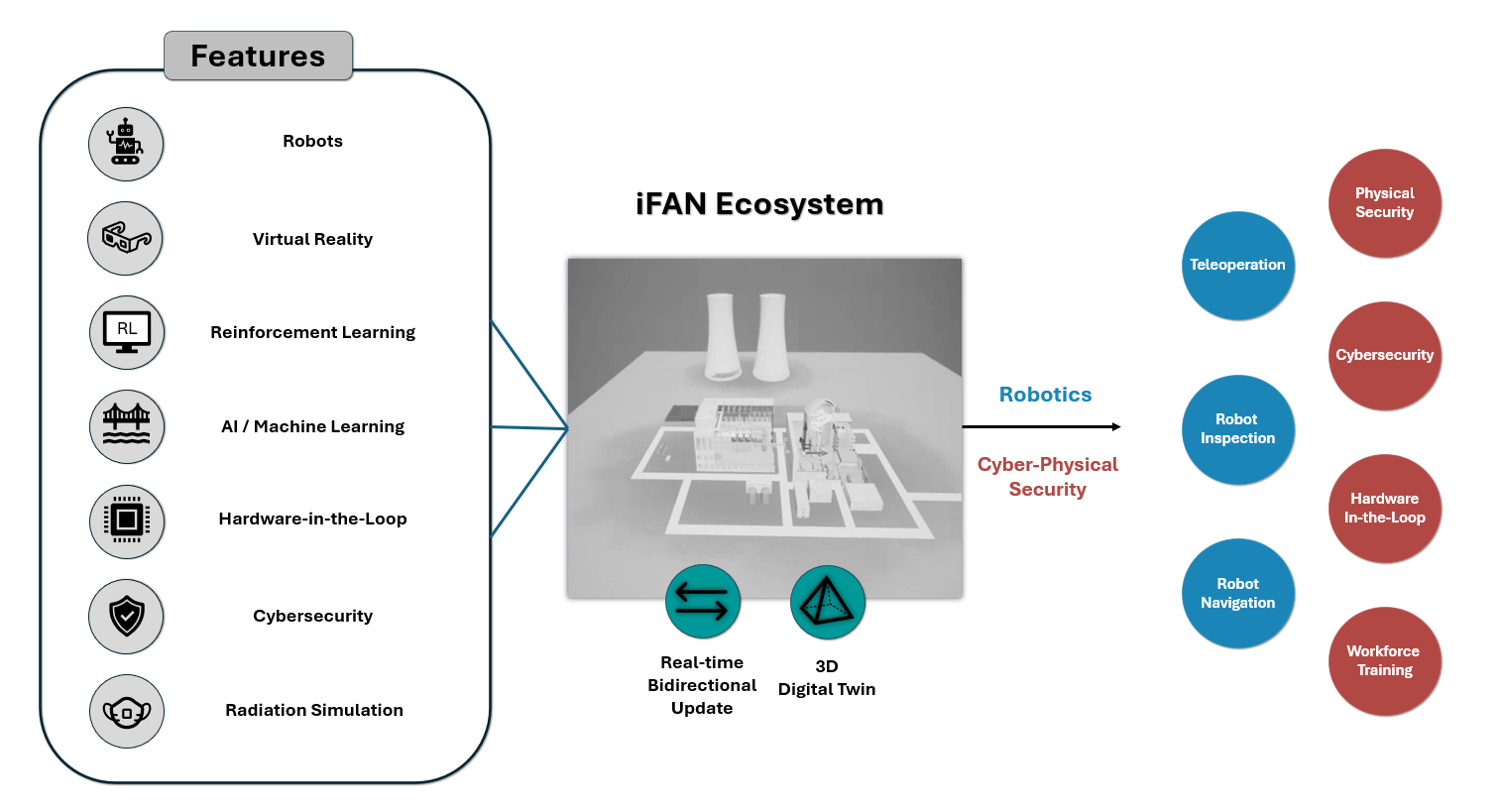}
    \caption{Architectural overview of the iFAN ecosystem. The diagram illustrates the integration of core features (left) such as AI, VR, and cybersecurity into a 3D Digital Twin environment (center). This framework integrates functionality pertaining to both robotics and cyber-physical security into a unified system, enabling capabilities such as robot inspection, teleoperation, and physical security testing (right).}
    \label{fig:iFANeco Overview}
\end{figure*}

Due to the broad nature of the iFAN ecosystem, this paper is divided into sections that best present the ecosystem's structure and numerous capabilities. Figure \ref{fig:iFANeco Overview} gives a visual representation of the ecosystem's components, listed in the areas surrounding the bird's eye view of the DT model. Some of these components include cybersecurity, autonomus control, and teleoperation, each of which is further discussed in this work. The DTs communication with GPWR is depicted with the smaller symbols inside of the main DT figure. Section \ref{sec:Related Works} discusses separate similar works, Section \ref{sec:3D Modeling} describes the constructed 3D model of the iFAN DT, Section \ref{sec:Core Features} presents the core features of the ecosystem and their individual underlying functionality, Section \ref{sec:Demonstration} describes ongoing research within the ecosystem that employs the listed core features, and Section \ref{sec:Conclusion} concludes and summarizes the paper.

\section{Related Works}
\label{sec:Related Works}
In recent years, the deployment of autonomous systems has gained attention and require the development of DTs that facilitate high fidelity pre-deployment testing in NPP scenarios. Frameworks such as iFANnpp \cite{do2026ifannpp} and the multimodal research by Baniqued et al. \cite{baniqued2024multimodal} have demonstrated the power of integrating physics engines, such as Unreal Engine 5 (UE5) and Unity, with real-time plant simulators. These platforms allow for the validation of multi-robot path planning and radiation-aware navigation in a risk-free virtual space. Beyond robot navigation, specialized DTs have been leveraged to explore complex decommissioning tasks \cite{benkrid2025digital, jharko2021aspects}, highlighting the critical role in autonomous robotic monitoring. Furthermore, the integration of additional modalities, such as haptic feedback \cite{tugal2023impact}, has been developed to bridge the gap between virtual simulation and real-world robotic operation. Despite these advancements in realism, these frameworks often ignore the underlying communication vulnerabilities and control-layer interdependencies that are critical in a modern, digitized NPP.

A core feature of DTs is their ability to model a physical system to enable scenario validation, serving as a testbed for realistic development and testing. There have been multiple efforts to create testbeds integrating digital Instrumentation and Control (I$\&$C) systems to strengthen cyber-physical security of NPPs. Research surrounding the Purdue PUR-1 reactor \cite{theos2022development} has utilized an I$\&$C testbed for real-time anomaly detection. ARCADE \cite{maccarone2023advanced} has been developed to investigate the impact of malicious cyber-attacks on reactor cooling and control logic. HINT-Sec \cite{nam2025hint} was developed as a nuclear I$\&$C testbed that integrates safety-grade controllers and software to evaluate realistic fail-safe protocols under cyber-attack scenarios. Additional works have modeled control system functionality pertaining to a wide range of nuclear sub-systems within their testbeds, including the reactor coolant system \cite{zhang2019multilayer}, pressurizer \cite{lee2019development}, and steam generator \cite{chen2024full}. I$\&$C nuclear testbeds have been used to simulate a number of attack scenarios for cyber-defense development, including man-in-the-middle (MITM) \cite{zhang2018industrial}, denial-of-service (DoS) \cite{zhang2021developing}, and false data injection (FDI) \cite{allison2020plc}.  

State-of-the-art digital twins, including iFANnpp \cite{do2026ifannpp}, provide high-fidelity environments for robotic testing, yet they often lack the complex cyber-physcial capabilities. In contrast, security-focused digital twins rarely incorporate the rest of operation, such as autonomous robot capabilities. This paper closes the loop between these distinct platforms by proposing an ecosystem designed for AI, cyber-physical security, and robot simulation for NPP operations that can support cybersecurity, physical security, operational fault detection and diagnostics, robotics operation, virtual reality, and workforce training.

\section{3D Modeling}
\label{sec:3D Modeling}

In order to give a proper testbed for experimentation and process development, the iFAN ecosystem utilizes its own DT that recreates a 3D interactive NPP. GPWR alone provides very limited spatial data, complicating a transition from a two dimensional to three dimensional space, so the model was developed with the same sequential layout and connections, but with full autonomy for relative height and locations. The resulting DT gives a playground for the implementation and assessment of the multiple facets found within the iFAN ecosystem. Having the 3D model gives spatial feedback to the robots and their tasks without the need for expensive construction costs or safety concerns in a real testing environment. The further inclusion of human and robot controllable characters gives the opportunity for training without the undesirable possibility of danger to the plant systems.

Previously, iFANnpp included baseline models divided into the following zones: the reactor building, the turbine hall, the water source and cooling towers, and the power distribution area  \cite{do2026ifannpp}. The included updates focus primarily on systems found in the reactor and turbine buildings, with the main goal being to further develop the model to match a GPWR more closely. Since the previous iFANnpp model, major systems involved in the steam cycle for nuclear power generation have been added and expanded upon. These include condensate, feedwater, pressurizing, and main steam systems. Additional modifications have been implemented on the reactor core and control rods which will be discussed later in this section. Although UE5 has the capability to import models, all components were created using existing geometric shape models from the Unreal software.

\begin{figure}[t]
    \centering
    \includegraphics[width=0.9\linewidth]{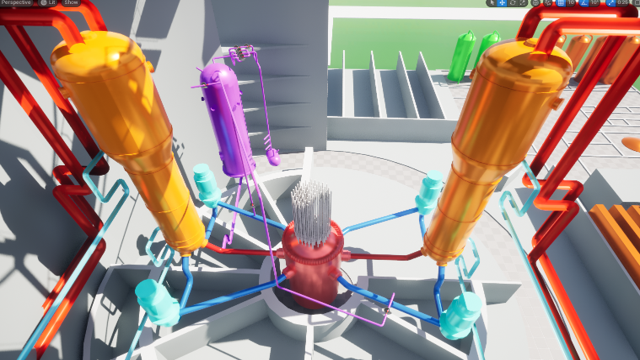}
    \caption{\centering Primary System Model}
    \label{fig:RCS}
\end{figure}

Since the pressurizing system had already been partially developed, this was the first focal point for improvements. Components added to the vessel were a spray valve, heaters, relief valves, and the relief tank. Each component has redundancy, interconnections, and control capabilities defined within GPWR. The overall system was given a purple color and connects directly to the main coolant system via a single surge line and two spray lines. The GPWR simulator also has a unique coolant system layout, which was the second system addressed. Most systems tend to have coolant loops consisting of a single steam generator (SG), a cold leg with one pump, the reactor core, and a hot leg returning to the SG. However, this system uses coolant loops consisting of two cold legs with one pump each, and one hot leg per SG, with each SG having two steam outlets and two feedwater inlets. Starting from the original design in the predecessor work, the reactor core and attached SGs were rotated, and the second cold legs were added. This led to a modification of the feed and steam piping to match the new configuration by using the modeling function within UE5. The final pressurizer and coolant loop 3D model can be seen in Figure \ref{fig:RCS}.

\begin{figure}[t]
    \centering
    \includegraphics[width=8cm,height=4.5cm]{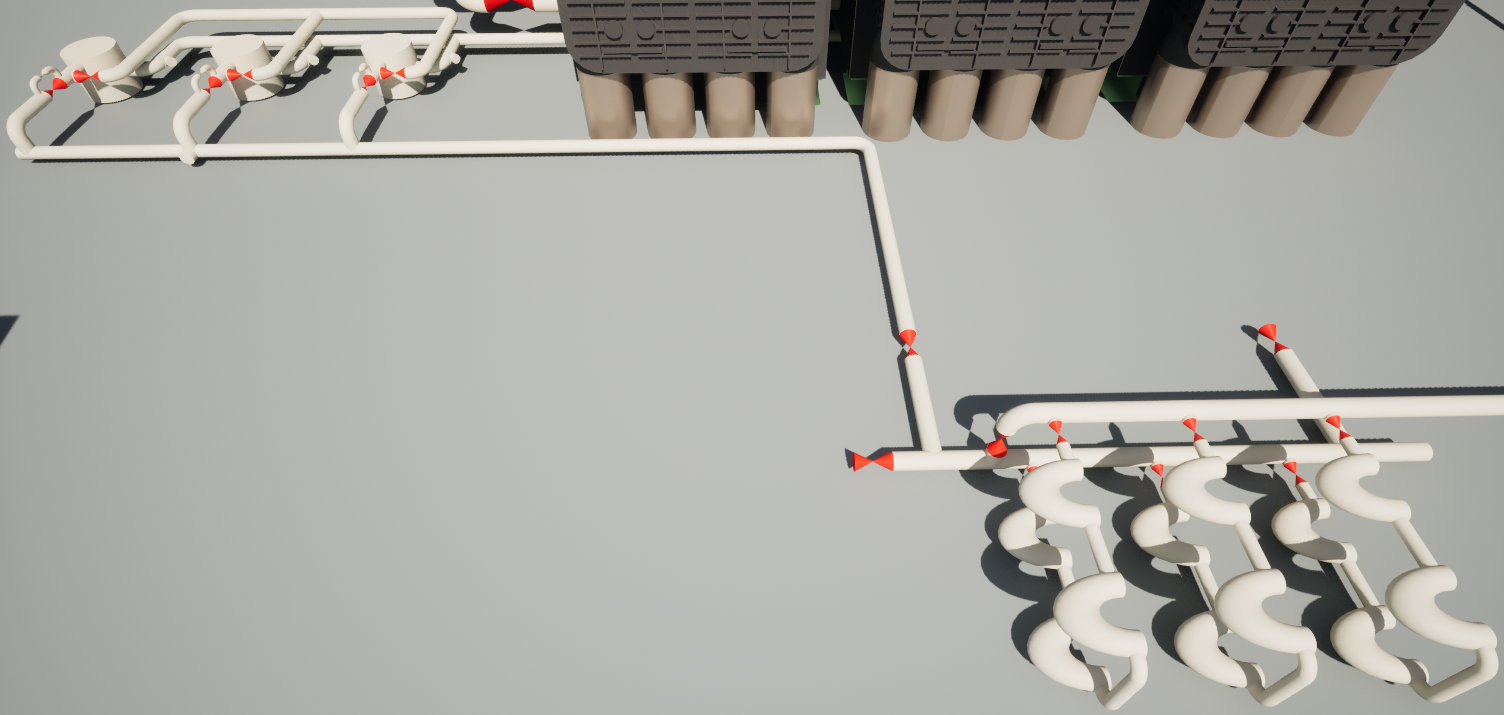}
    \caption{\centering Condensate System Model}
    \label{fig:COND}
\end{figure}

After the coolant and pressurizing systems were adjusted, the secondary steam systems were developed, starting with the condensate system. This model is split into two sections: the pumps and the heat exchangers. The pumps, which are modeled with a cylinder, are connected in parallel to the main condensers and discharge to a common header towards the heat exchangers. Recirculation lines for each pump and a cutoff for connection to a yet-to-be modeled demineralizer are included as well. In the next stage, there are three parallel sets of four series-connected heat exchangers to help preheat the condensate with isolation valves in each set and an overall bypass. The three pathways all discharge again to a common header that leads to the main feedwater system. The entire connected condensate system is shown in Figure \ref{fig:COND}.

\begin{figure}[t]
    \centering
    \includegraphics[width=0.9\linewidth]{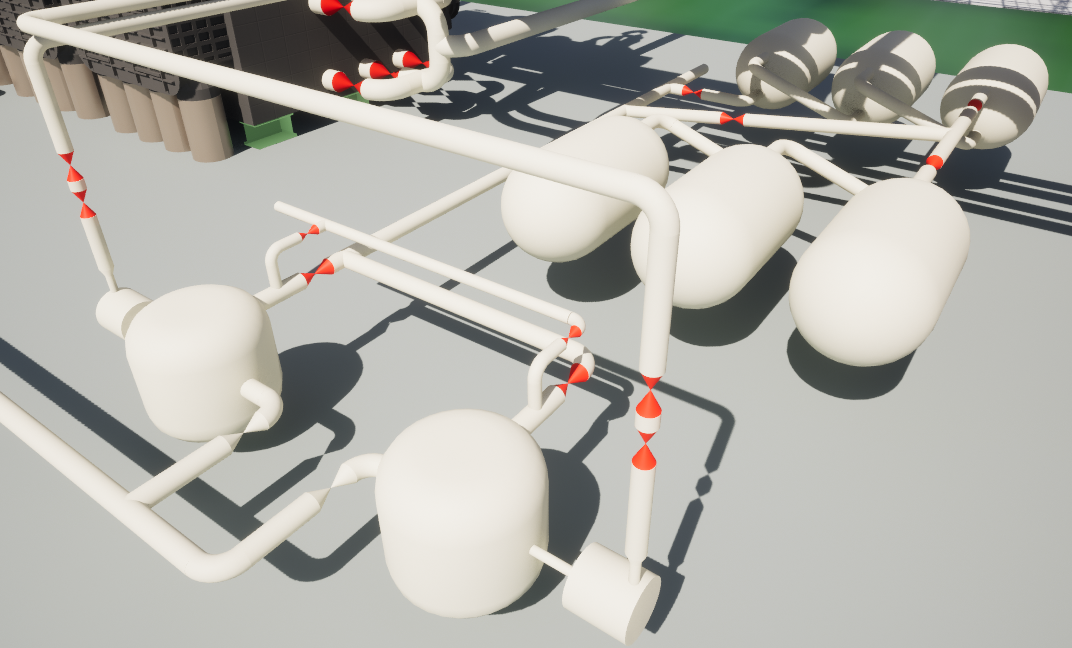}
    \caption{\centering Main Feedwater System Model}
    \label{fig:MFW}
\end{figure}

The main feedwater system utilizes two parallel pathways starting with the combined suction of two turbine-driven feed pumps from the discharge of the condensate system. These two pumps discharge to a combined pipe that again splits to two parallel sets of three series connected heat exchangers. These final two pathways merge and then split once more into the final feed pipe to each SG. This pipe was modified from the original version to include the water level control valves and to match the two SG design. The final feedwater system is shown by Figure \ref{fig:MFW}.

\begin{figure}[b]
    \centering
    \includegraphics[width=0.9\linewidth]{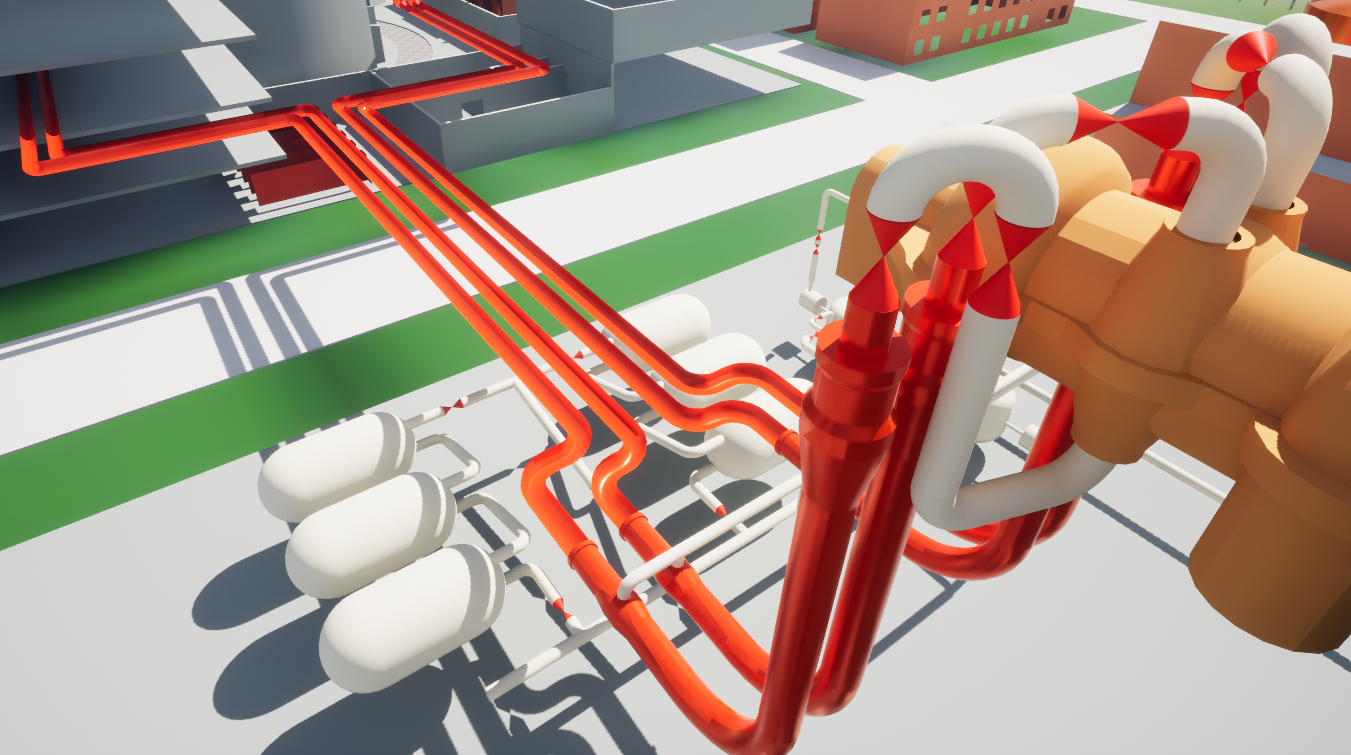}
    \caption{\centering Main and Reheat Steam System Model}
    \label{fig:MRS}
\end{figure}

To complete the basic steam cycle, the current DT includes further developments to the main steam system. As mentioned, the piping from the SGs was reconfigured to have two outlets per SG, shown by the four main pipes in Figure \ref{fig:MRS}. Further additions include the main steam stop valves, connections to the high pressure turbine, and piping to supply the feed pump turbines. The main piping is red for system clarification, and white represents the subsequent system branches.

In reactor core, the control rod modeling is set up as a 17x17 grid, matching the GPWR simulator used in this enviroment. To open more testing avenues, the iFAN ecosystem 3D model needed to include interaction capability. This ability is desirable for training both human and robotic operators as it eliminates the risk of damaging existing physical systems and personnel. The added novelty of the iFAN ecosystem is that the DT not only models a pressurized water reactor, but it also shares data with GPWR via a bidirectional Python-based bridge to provide real-time updates to changes in plant status. GPWR has a local user interface (UI) that allows operation of components and implementation of pre-programmed malfunctions.

\begin{figure}[t]
    \centering
    \begin{tabular}{cc}
    \includegraphics[width=0.48\linewidth]{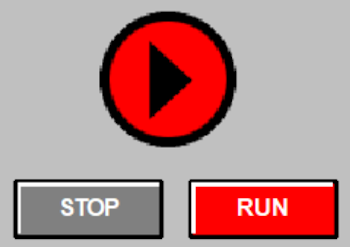} &
    \includegraphics[width=0.48\linewidth]{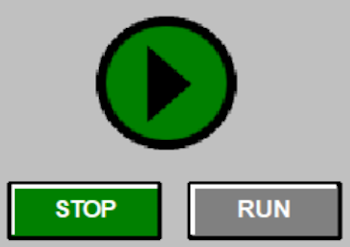} \\
    \includegraphics[width=0.48\linewidth]{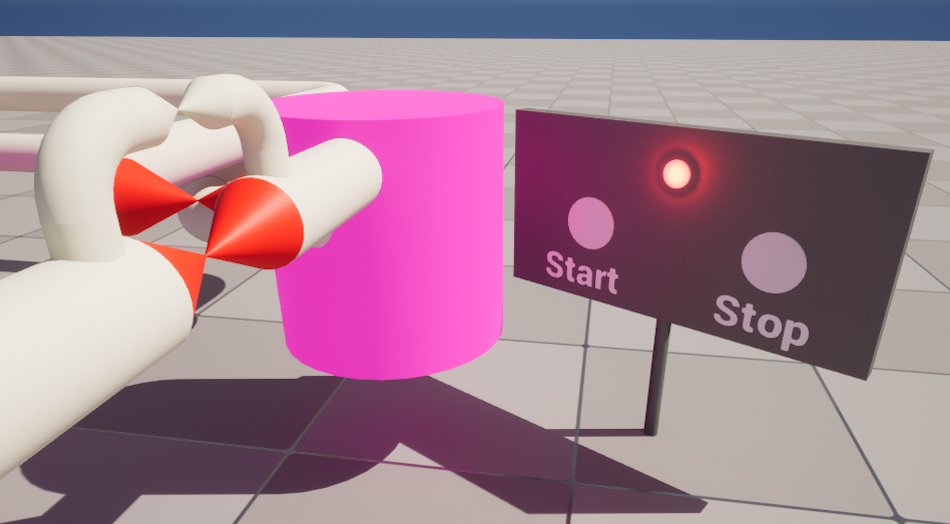} &
    \includegraphics[width=0.48\linewidth]{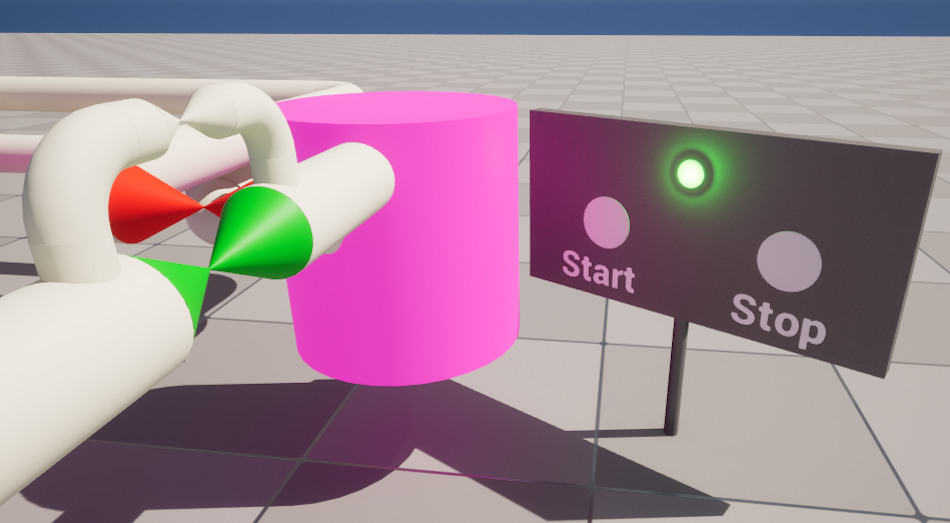}
    \end{tabular}
    \caption{\centering Component Status Scheme}
    \label{fig:Component Status}
\end{figure}

For indication to the operator, the DT was set up again to match the output of GPWR . The currently integrated indications are shown in Figure \ref{fig:Component Status}. Red indicates running or open components, while green indicates secured or shut components, such as a valve. The pink coloration of the component shown is a visual representation of temperature that uses a blue to red gradient transition, blue being colder. 

\section{Core Features}
\label{sec:Core Features}
The iFAN ecosystem has integrated a set of core features that allows multiple personnel to conduct focused research and development on realistic cybernetic, cyber-physical, and robotics systems within the context of a NPP. This provides an integrated safe and effective testbed for cybernetic systems and robotics. The aforementioned features focus on frameworks in virtual reality (VR), reinforcement learning (RL), AI and machine learning (ML), cybersecurity, radiation simulation, and hardware-in-the-loop (HIL) capabilities. Each of the following subsections describes these features in detail.

\subsection{Virtual Reality}

An important focus in the nuclear field is to enhance safety in the performance of both common and complex operations. VR provides operators with a safe, effective, and translatable environment to achieve desired outcomes. UE5 provides native support for Meta Quest 3 through the OpenXR and Meta XR plugins, which gives a stable foundation for VR interaction. On top of this, the iFAN ecosystem adds three preset VR modes that make it easier to configure different VR behaviors depending on the user's desired operation.

The first mode, \textit{BP\_Human\_FirstPerspective}, aligns a camera with the head of a basic mannequin included in the baseline UE5 software. Because the camera directly follows the trainee’s headset, this mode provides a natural first-person view of the environment. It is primarily used for training scenarios where spatial awareness, hand–eye coordination, and close-range interaction are important. The same VR widget system allows the user to select buttons, open and close valves, interact with UIs, and manipulate any objects. These actions can be accomplished in both an open world format for the entire NPP or through a predefined scenario if a specific task is desired.

In addition to the human-centered VR mode, two modes are designed specifically for robot teleoperation. The First-Person Teleoperation Mode places the camera on the robot’s camera frame. The operator views the environment as the robot does, which is useful for tasks requiring  fine manipulation and is an effective means of testing teleoperation capabilities before real-world deployment. Inputs from the VR controllers are mapped to the end-effector actions, and the robot's response in the DT is shown in real time.

Similarly, the final Third-Person Teleoperation Mode positions the camera behind or slightly above the robot. This view makes it easier to judge global surroundings, avoid collisions, and plan navigation paths. The camera offset can be fixed or adjustable depending on the task. Compared to the first-person robot mode, this configuration trades immersion for better situational awareness. In this mode, control inputs from the VR controllers are mapped to the robot movement actions.

All three are kept as independent GameMode templates so that switching between them does not require modification of the underlying \textit{VR Pawn} or controller blueprints. This structure is advantageous in that it streamlines demonstrations, behavior documentation, and maintains consistent input mappings across various definable scenarios and teleoperation procedures. In this way, the VR capability of the iFAN ecosystem directly enhances training and development of nuclear operators.

\subsection{Reinforcement Learning}

Another aspect of the iFAN ecosystem is its ability to support RL-based model training, which allows for the development of learning-based control and design algorithms. Learning-based methods aim to produce efficient control algorithms by experimentally adapting from data, providing the possibility to handle complex dynamic scenarios that may be difficult to implement with manual rules. To facilitate the use of the ecosystem for RL, a Python-based RL environment is implemented.

The RL environment implements the Gymnasium API \cite{towers2024gymnasium}. This environment standard is chosen due to its compatibility with many reinforcement learning frameworks, such as Stable Baselines3 \cite{stable-baselines3}, RLib \cite{pmlr-v80-liang18b}, and TorchRL \cite{ICLR2024_07bc8125}, supporting portability between projects utilizing different RL libraries. All API methods \textit{step}, \textit{reset}, \textit{render}, and \textit{close} are implemented to enable model training for autonomous robot control. 

Communication between the RL environment and the iFAN ecosystem environment is enabled by the UnrealCV plugin \cite{qiu2017unrealcv}. Communication is implemented with a client-server architecture. A C++-based server integrated with the iFAN ecosystem's UE5 DT environment can interact with all components within the level, allowing for reading and modifying UE5 actor fields, as well as controlling actor actions like movement, and environment interactions, such as button presses. The Python-based client is integrated with the provided RL environment, which acts as the intermediary between the Python-based RL learning algorithm and the DT environment.

\begin{figure}[t]
    \centering
    \includegraphics[width=\linewidth]{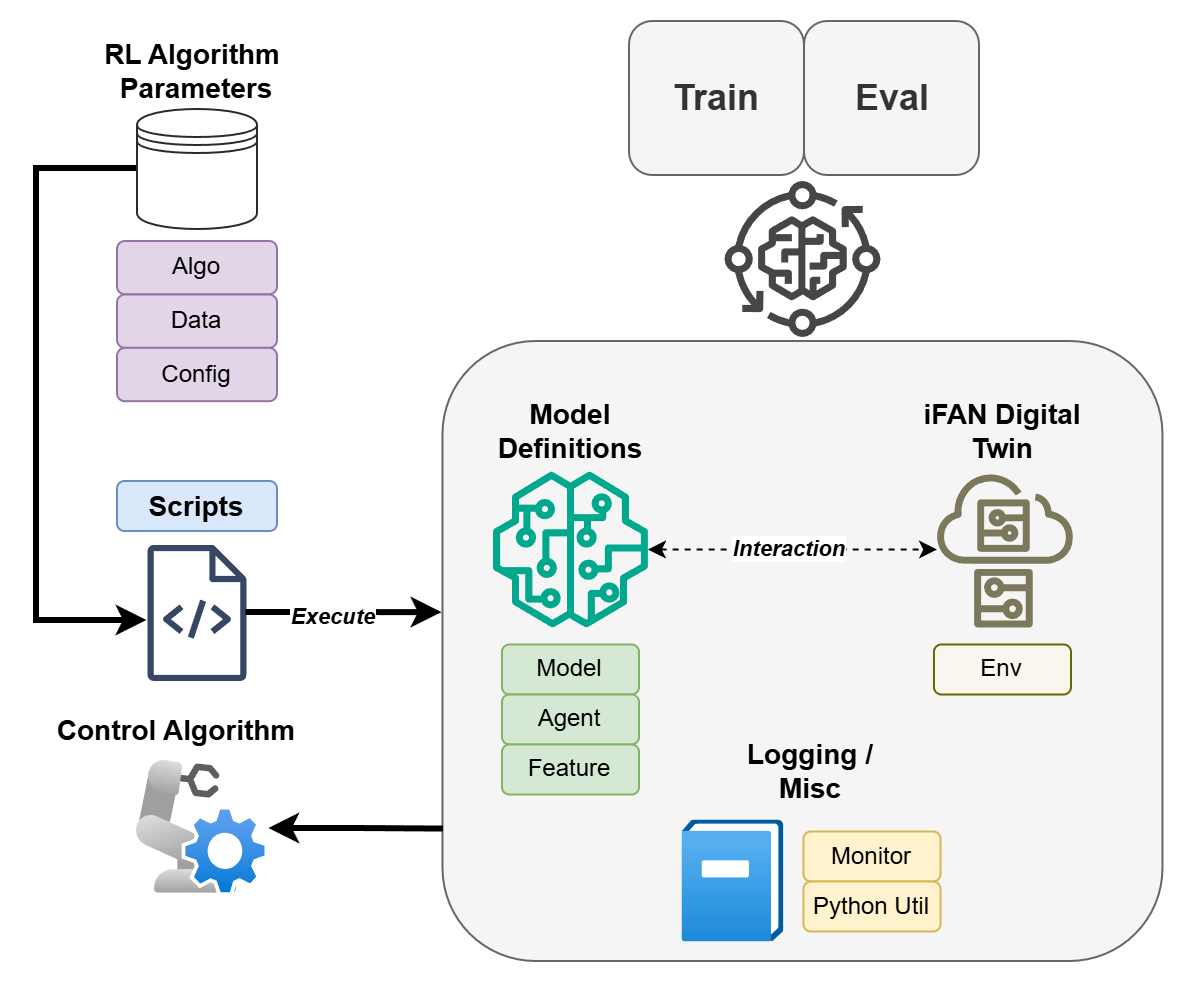}
    \caption{\centering RL Project Template Functional Diagram}
    \label{fig: RL-template-pipeline}
\end{figure}

State information about a moving robot is queried from the iFAN DT, giving robot location, goal location, as well as collision information as needed. Additionally, the movement of the robot can be separately controlled within the RL environment. The provided RL environment also demonstrates both data querying and actor control when interacting with the iFAN DT. This functionality is implemented by defining commands within the UnrealCV C++ server \textit{ObjectHandler} object, which allows for defining functions that interact with the digital NPP. A command name and a set of parameters are mapped to a C++ function that runs the game engine code, and returns a response as defined by the developer. The Python-client can then send this pre-defined command to the server, and additionally incorporate the response into the RL training. Modifications to our provided RL environment can be made to support additional tasks by adding additional functions to the UnrealCV C++ server and executing them with the Python client. The example RL environment demonstrates this generic workflow that allows for modifying actor (e.g., robot) controls, and extending the state information that can be queried from the iFAN DT to facilitate end-to-end interaction and learning.  

The RL environment is contained within a generic directory structure serving as an RL project template, designed to train and evaluate reinforcement learning algorithms through interaction with the iFAN DT. The project template can be used to define and extend RL learning projects within the iFAN ecosystem, providing a standardized structure for project development. Figure \ref{fig: RL-template-pipeline} displays the functional components within the RL project template, which are defined below.

\begin{itemize}
    \item \textbf{RL Algorithm Parameters}: RL algorithm implementations (under algo), algorithm configuration files (under config), and potential offline data for training or evaluation (under data). These are specified within the launch scripts.
\end{itemize}
\begin{itemize}
    \item \textbf{Scripts}: Programmatic scripts used to execute train or evaluation code with predefined parameters, to streamline running different experiments.
\end{itemize}
\begin{itemize}
    \item \textbf{Train / Eval}: Directories containing scripts for executing generic training and evaluation loops respectively. 
\end{itemize}
\begin{itemize}
    \item \textbf{Model Definitions}: Logic relevant to the control algorithm being trained and evaluated. Includes trainable model architecture definitions (under models), code to wrap the underlying controller model and handle model input and output (under agent), and potential feature extraction or preprocessing (under feature). 
\end{itemize}
\begin{itemize}
    \item \textbf{iFAN DT}: Environment directory containing environment definitions (e.g., following the gymansium API), used with the control algorithm to interact with iFAN DT environment during training.
\end{itemize}
\begin{itemize}
    \item \textbf{Logging / Misc}: Miscellaneous utility functions that can be created and defined throughout the project (under python util) as well as custom logging logic to track loss values or performance metrics during training or evaluation (under monitor).  
\end{itemize}

Within the feature, each functional component is split into its own directory, each denoted in the figure with a colored box. The figure groups the components based on their relationships within the RL training and evaluation pipeline. A full demonstration of the functionality of the RL environment is discussed in Section \ref{sec:Robot Navigation} using a robot navigation example. The layout of this RL framework serves as a foundational baseline upon which future RL techniques can be addressed, where physical interaction with the plant during training is enabled and where the framework can support further developments in automated robotic maintenance and plant design optimization.

\subsection{AI $\&$ ML}

The iFAN ecosystem also includes AI and ML as a core feature. GPWR generates physics-based, realistic simulations under various scenarios. This produces datasets that reflect how a real system behaves, including sensor readings and system responses. This makes data collection safe and repeatable. The main goal of this module is to support a “data → model → validation” loop. After the data is extracted from the DT and preprocessed, users can run common ML models directly on it. After training, the model can be deployed back into the DT pipeline in the iFAN ecosystem and executed inside simulated scenarios. This closed-loop validation will speed up AI/ML deployment in NPPs.

\begin{figure}[t]
    \centering
    \includegraphics[width=\linewidth]{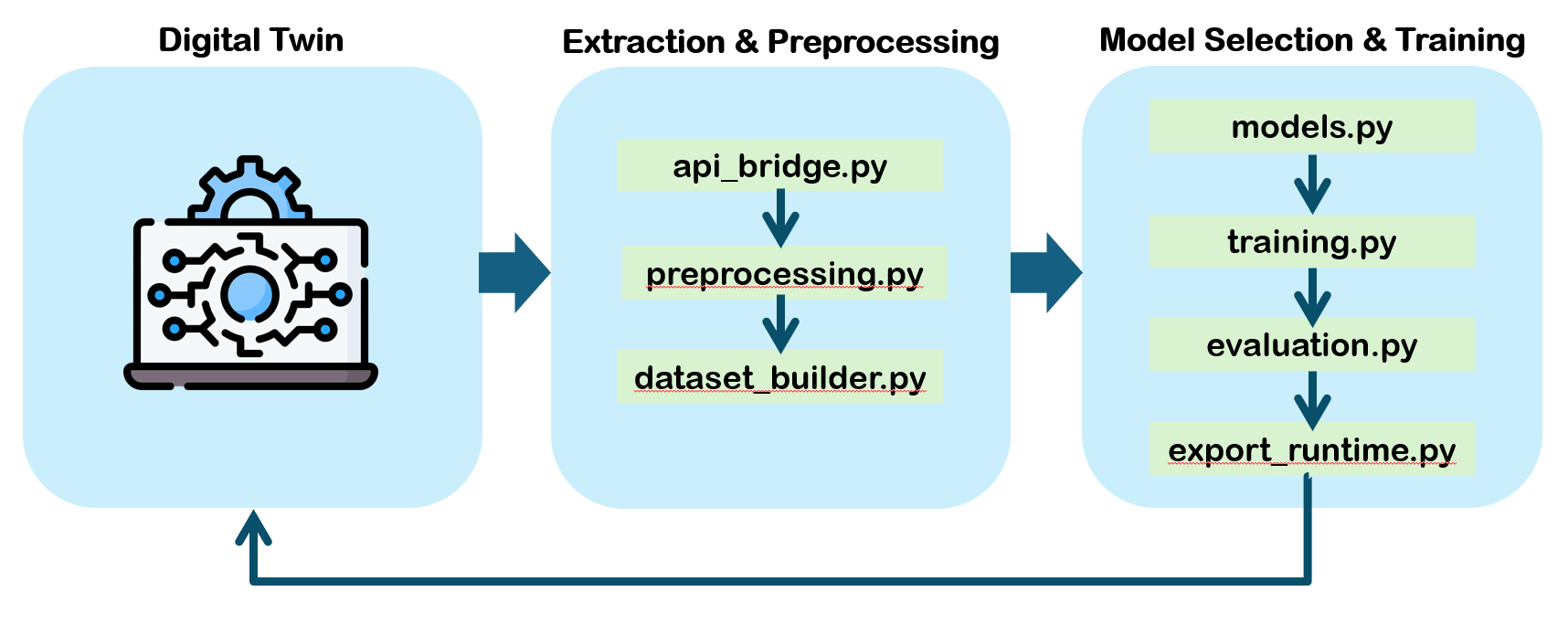}
    \caption{\centering Data → Model → Validation pipeline for AI/ML deployment in iFAN ecosystem }
    \label{fig: AI/ML pipeline}
\end{figure}

With this pipeline, the iFAN ecosystem brings two major benefits. First, iFAN provides a reusable pipeline that covers data capture, data processing, model training, evaluation, and deployment, which are shown in Figure \ref{fig: AI/ML pipeline}. This pipeline reduces engineering cost and makes results easier to repeat. The same pipeline can be reused across different tasks and scenarios without changing the main workflow.

In the data extraction step, data is first pulled through platform interfaces that connect user code to GPWR. A common setup is a Python API bridge (\textit{api\_bridge.py}) that allows experiment scripts to read sensor output, query environment or equipment states, and collect logs in real time. This design ensures that the training and testing code can interact with the simulator step by step and obtain the state feedback for learning, prediction, or decision-making.

After data extraction, the pipeline processes raw outputs into model-ready inputs, which is done by \textit{preprocessing.py}. This step will format the sensor signals into vectors or tensors, normalizes and encodes state variables, or constructs labels. The processed outputs are then packaged by \textit{dataset\_builder.py}, which constructs a reusable dataset bundle. The pipeline supports common dataset management functions, such as train, validation, and test split, and can optionally include data augmentation to improve model robustness.

Following the preprocessing step, the model is selected and trained. The various types of models can be used to analyze the preprocessed dataset. These include time-series models for forecasting and anomaly detection, as well as classical ML methods such as regression, classification, and clustering. The setups of candidate model are organized in \textit{models.py}, and optimization and training routines are implemented in \textit{training.py}. Model performance is checked in \textit{evaluation.py}. After training, the selected model is exported in a standard format via \textit{export\_runtime.py} and then reconfigured for closed-loop testing and validation in the ecosystem.

Another advantage of the AI and ML aspect in the iFAN ecosystem is that evaluation can be performed under the same conditions by reseting the GPWR to replay the scenarios and measuring the metrics such as accuracy, false alarm rate, and success rate, which can support a fair comparison between different model. Besides, the robust of model can be evaluated by adding noise or introducing sensor faults. Visual outputs and logs from GPWR provide additional support for debugging and behavior review. Once a model is validated, it can be deployed back into the iFAN DT pipeline, in which model reads sensors and produces predictions or decisions, and GPWR then applies these outputs to the environment or control interface. The resulting closed loop allows end-to-end testing before moving to real systems. Finally, real operational data is fed back to enrich the dataset and trigger further training, forming a simulator-to-real NPP iteration loop that improves robustness.

\subsection{Cybersecurity}
\label{sec:Cybersecurity}

\begin{figure*}[t]
    \centering
    \includegraphics[width=0.6\textwidth]{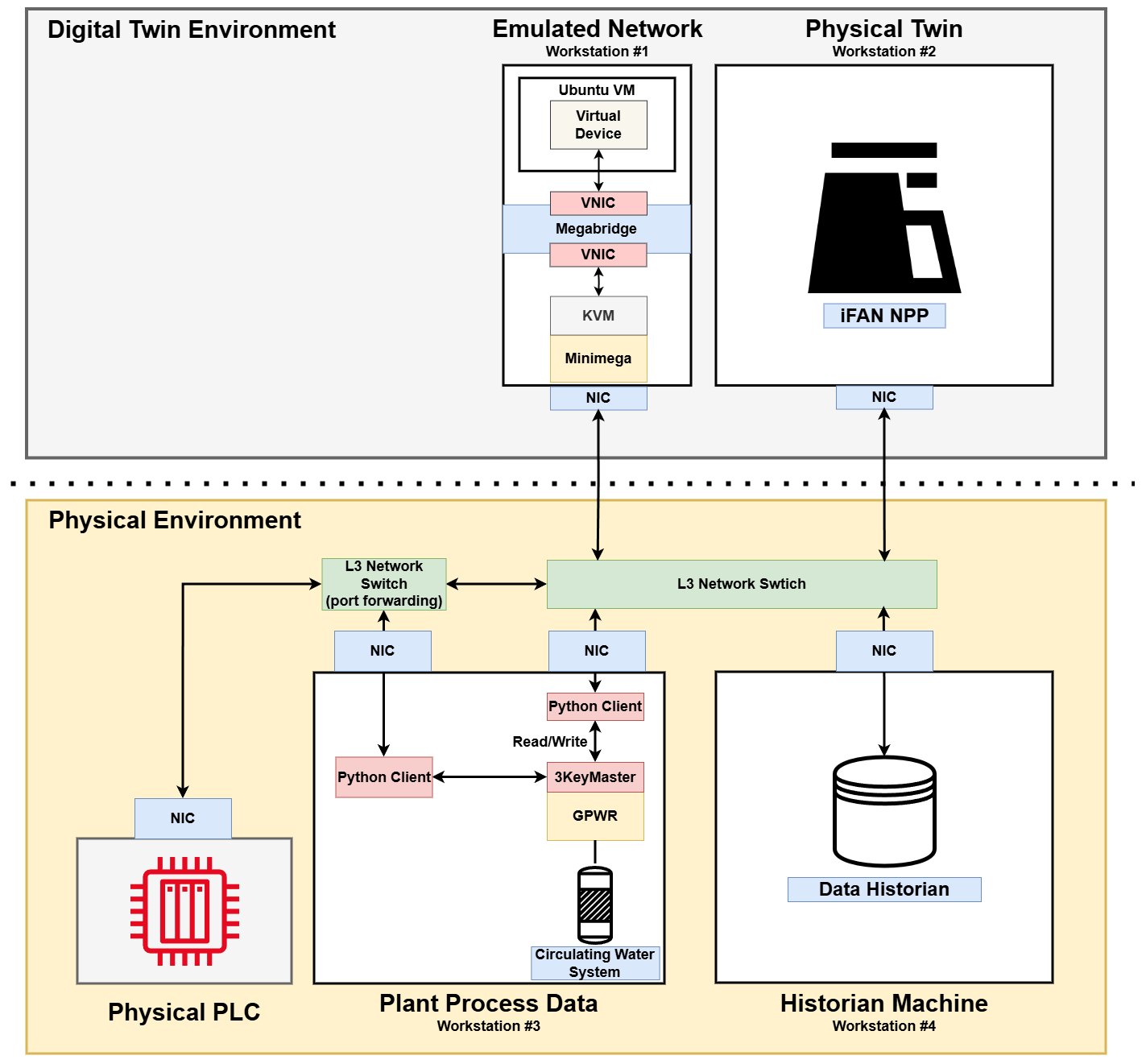}
    \caption{Network Diagram and Overview of Physical and Digital Twin Environment}
    \label{fig:ndt_ecosystem}
\end{figure*}

Further capability of the iFAN ecosystem includes its ability to integrate with a physical networked testbed to support synchronization with live data collected through the network. This enables cyber-attack scenarios and mitigation strategy development that considers both network-based and physical effects. Figure \ref{fig:ndt_ecosystem} displays the testbed architecture. The testbed integrates a physical programmable logic controller (PLC), a simulation workstation, and a data historian to emulate a physical industrial nuclear environment. Additionally, the iFAN ecosystem as well as an emulated network serve as mirrored virtual representation of the physical systems. The components are networked together through a Layer 3 switch, forming the primary network. 

The NPP is simulated on workstation \#3, which is an Intel(R) Xeon(R) W-2245 machine containing 8 cores and 16 threads. This machine is the primary host for the GPWR simulator, which provides the realistic plant process data that is propagated through the testbed network. A PLC from the circulating water system is modeled with a physical Allen Bradley control Logix 5580 1756-L81E PLC within the testbed. The PLC tags are updated with sensor values read from GPWR, and written to the PLC using a Python client through a separate switch, to simulate the PLC reading from physical sensors. Workstation \#4 contains a data historian implemented with InfluxDB, which reads the physical PLC tags through the primary network and stores them over time. Within the testbed, this is implemented using an Intel(R) Xeon(R) W-11855M machine containing 6 cores and 12 threads. Process variable data are read from the data historian into the iFAN ecosystem 3D model, running on workstation \#2, which is a Intel(R) Xeon(R) w7-2475X machine with 20 cores and 40 threads. Additionally, an emulated network implemented with Minimega \cite{minimega} runs on workstation \#1, which contains a network that matches the topology of the physical network, with physical machines represented by virtual machines.

The iFAN ecosystem environment supports networked integration with the testbed for ingesting process variable data from the data historian. This allows for streaming live data from the physical testbed, which can subsequently update the physically modeled NPP within iFAN ecosystem, to enable cyber-attack scenario and mitigation strategy development that considers both network-based and physical effects. To support this, a networked sensor manager object was created to allow for polling the most recent process variable values from the testbed. The object monitors a list of process variables, specified from a configurable CSV file, and polls them from the data historian. The networked sensor manager object can then be added to a UE5 level, and the live sensor values streamed from the data historian can be utilized within the iFAN ecosystem. In Section \ref{sec:Demonstration}-F, a simulated cyber-attack is conducted to showcase this framework. 

\subsection{Radiation Simulation}
Since the iFAN ecosystem is based inside of the UE5, researchers have the capability to create custom functions within the UE5 software. The previous work \cite{do2024radiation} used OpenMC \cite{ROMANO201590} and UE5 together to demonstrate radiation flux visualization in the 3D environment. Further development has been conducted to execute and convert flux values into dose rates with Monte Carlo simulations. Finally, the collected data is imported to give an accurate radiation map in the 3D environment. This is accomplished through three custom parts: a new data table type, a dosimeter scene component, and a radiation source actor.

\textit{Data table} is a feature in UE5, which allows users to import text or Comma-Separated Value (CSV) files and to give the columns specific labels without having to parse them for features every call. In this instance, a custom data table labeled \textit{dose table} uses data generated from OpenMC, employing a regular 3D mesh to create a list of 3D voxels. The dose table consists of six columns: one for row name, three for the indices of the mesh, and the final two for dose rate information given in Sieverts per second (Sv/s) and per hour (Sv/hr).


The dosimeter is designed as a scene component to be attached to any actor type. This setup allows for complex systems, such as robots or DT pawns, to carry the dosimeter anywhere they travel. The dosimeter itself has two parts: the function to retrieve and update the dose and dose rate variables, and the widget updater to allow for an on-screen display of both the cumulative dose and the dose rate Sv/hr. This tandem is designed to mimic real dosimeters used by workers in a NPP, where they are defined as the maximum dose and dose rates given on approved work permits. The dosimeter update function calls the update dose function from the radiation source actor. The resulting dose rate outputs are converted from Sv/s to a cumulative dose value, and both outputs are then fed into a display widget. The widget uses the two values, reduces them into scientific notation, and displays the final cumulative dose and the Sv/hr dose rate on the screen of the simulation.

\begin{figure*}[t]
    \centering
    \includegraphics[width=\textwidth]{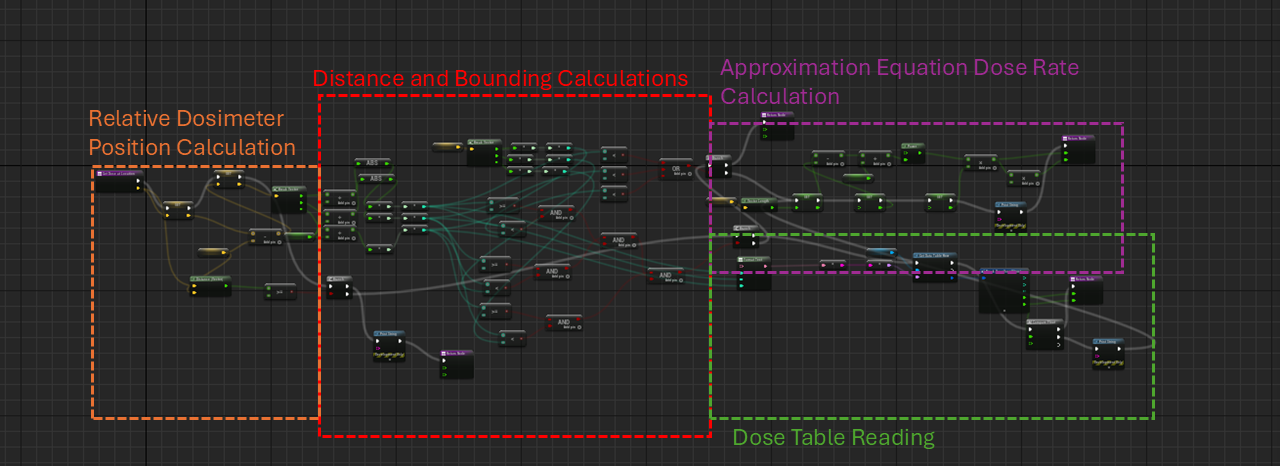}
    \caption{Update Dose Function of Radiation Source Actor}
    \label{fig:update_dose}
    \vspace{-1em}
\end{figure*}

The radiation source actor was developed to have features easily editable by the end user, namely the dimensions, voxel size, and origin point. It also houses the previously described function to update the dose for the dosimeter. This function can be dissected into four key parts as seen in Figure \ref{fig:update_dose}: the relative dosimeter position calculation, the distance and bounding calculation, the dose table reading, and the approximation equation dose rate calculations. The relative position calculation determines the relative position of the dosimeter from the source position found in the indices list. The distance and bounding calculation determines the distance between the dosimeter and source, and determines if the dosimeter is classified as in bounds, out of bounds of the dose table indices but within the bounds of the approximation equation, or fully out of bounds, in which case a zero will be returned. The dose table reading uses the index from the distance calculation, reads the rows of the dose table to find the i, j, and k indices of that voxel, and then returns the dose rates of that row. The approximation calculation occurs regardless of whether the dosimeter is further away than the Monte Carlo simulation table's range. Therefore, if the dose table read returns a zero, or if the simulation returns a zero where there should not be, the function is still called. The final dose rate is based on the exponential decay equation for gamma radiation, shown by Equation 1.

\begin{equation}
    \dot{D}(x) = \dot{D}(x_0)\, 0.5^{(x - x_0)/L}
\end{equation}

The equation is formatted in a halving distance format, where $\dot{D}(x_0)$ is the dose rate at the edge of the regular mesh used for baseline dose rate, $(x - x_0)$ represents the distance from the source, and $L$ is the halving distance of the dose rate. The equation outputs $\dot{D}(x)$ in Sv/s, and there is an included conversion to Sv/hr for the dosimeter. Since both the source and the factors of the equation can be configured to suit the user's needs, the iFAN ecosystem can support unlimited variations for both human and robotic scenario testing.

\subsection{Hardware-in-the-loop}
In conjunction with the GPWR simulator, an automation HIL testbed is deployed in the form of PLC equipment typically used in industrial control systems (ICS). The HIL testbed works by replacing functionality found in the simulator, such that its logic and programming can be managed by external real-world ICS hardware. This architecture keeps the rest of the NPP physics and control environments inside GPWR, while moving one or more functions into the scope of external control hardware with physical data exchange shared via a network switch. To communicate with the PLC, GPWR calculates and provides real-time industrial process variables (levels, flows, pump speeds, valve positions, etc.), as well as discrete signals (status inputs, operator commands, run indications, etc.) via an open platform communications (OPC) server, FactoryTalk Linx. This data can be read into the program-level tags found in the PLC’s memory space. The PLC then makes decisions and computations similar to GPWR, but it does so in the form of function block programming used for ICS.

The HIL testbed performs runtime execution based on the variables and signals found outside of its primary function and sends the resulting process variable changes and actuator commands back through the OPC, which effectively instructs GPWR on how to manipulate associated actuator outputs as well as provide new data for the simulator to use. This preserves a realistic cyber‑physical loop where GPWR handles most systems, while ICS hardware closes the control loop on important process variables for the designed function.

To keep the hybrid system physically meaningful, it is necessary to implement explicit simulator-to-PLC synchronization logic, with timing that provides for identical outcomes whether the HIL is in control of its functions or all functions are being managed by GPWR. The benefit of an automation HIL testbed is that the architecture combines the training‑grade, full‑scope, high‑fidelity GPWR simulator with real industrial hardware and industrial protocols, enabling detail-oriented experimentation for the purpose of cybersecurity research.

GPWR can simulate normal, transient, design basis accidents and customized malfunction conditions, so the impact of cybersecurity attacks on plant‑level safety margins and dynamics can be studied under realistic scenarios. One such study is shown in Section \ref{sec:Demonstration} with a focus on the steam generator (SG) level control system. The use of industrial-grade PLC hardware and industrial networking exposes genuine device and protocol behaviors and vulnerabilities, enabling quality research on supply‑chain compromises, firmware or logic tampering, PLC false data injection, as well as process‑aware attacks that depend on controller implementation details. It is for these reasons that HIL functionality has been integrated as an available core feature for the iFAN ecosystem.

\section{Demonstration}
\label{sec:Demonstration}
In this section, the previously presented core features are used to showcase the wide range of capabilities of the iFAN ecosystem in different areas of research, with a specific focus on the development of robotic and cybersecurity technologies for NPPs. These demonstrations often combine multiple core features available in the ecosystem and are underpinned by a foundational basis that demonstrates the versatility of a fully integrated NPP DT.

\subsection{Workforce Training}

Workforce training is an important function of the ecosystem that equips trainees with the ability to remember various operational activities and procedures, and to recognize and evaluate work-related risks \cite{norris2019virtual}. By using the ecosystem, trainees can not only become familiar with their future work environment but also retain the responses required under specific operational conditions. Compared to traditional educational methods based on textbooks, presentations, and video instructions \cite{zhao2023comparison}, the VR environment integrated with the features from the iFAN ecosystem has demonstrated effectiveness and efficiency in training system operators for nuclear power plants, where training on actual equipment or within a real environment is prohibited or very expensive. The current version of the ecosystem has been verified for training field operators inside NPPs \cite{do2024vrtraining}. The ecosystem can provide trainees with an immersive experience and help them understand their working environment.

Since the ecosystem integrates a 3D NPP model with the GPWR simulator, the ecosystem is able to reproduce the responses of physical components within the power plant. This integration can reproduce various operational scenarios in the NPP. Also, the ecosystem provides a user-friendly chatbot that delivers real-time text-based guidance to trainees. During each training session, the chatbot dynamically interact with the trainees to provide guidance and improve their task performance. After each training session, a detailed performance report is automatically generated, enabling trainees to objectively assess their interactions and identify potential areas for improvement.

The ecosystem offers several advantages over other VR training platforms \cite{xie2021review}. First, the training chatbot integrated into the ecosystem can interact with trainees, significantly improving the efficiency of training sessions. Also, the report system can directly elicit important information from the 3D environment, including task completion time, error types, and various response actions performed by trainees, so that the generated reports accurately reflect trainees' insufficiency and lack of knowledge.

\begin{figure*}[t]
    \centering
    \includegraphics[width=0.8\linewidth]{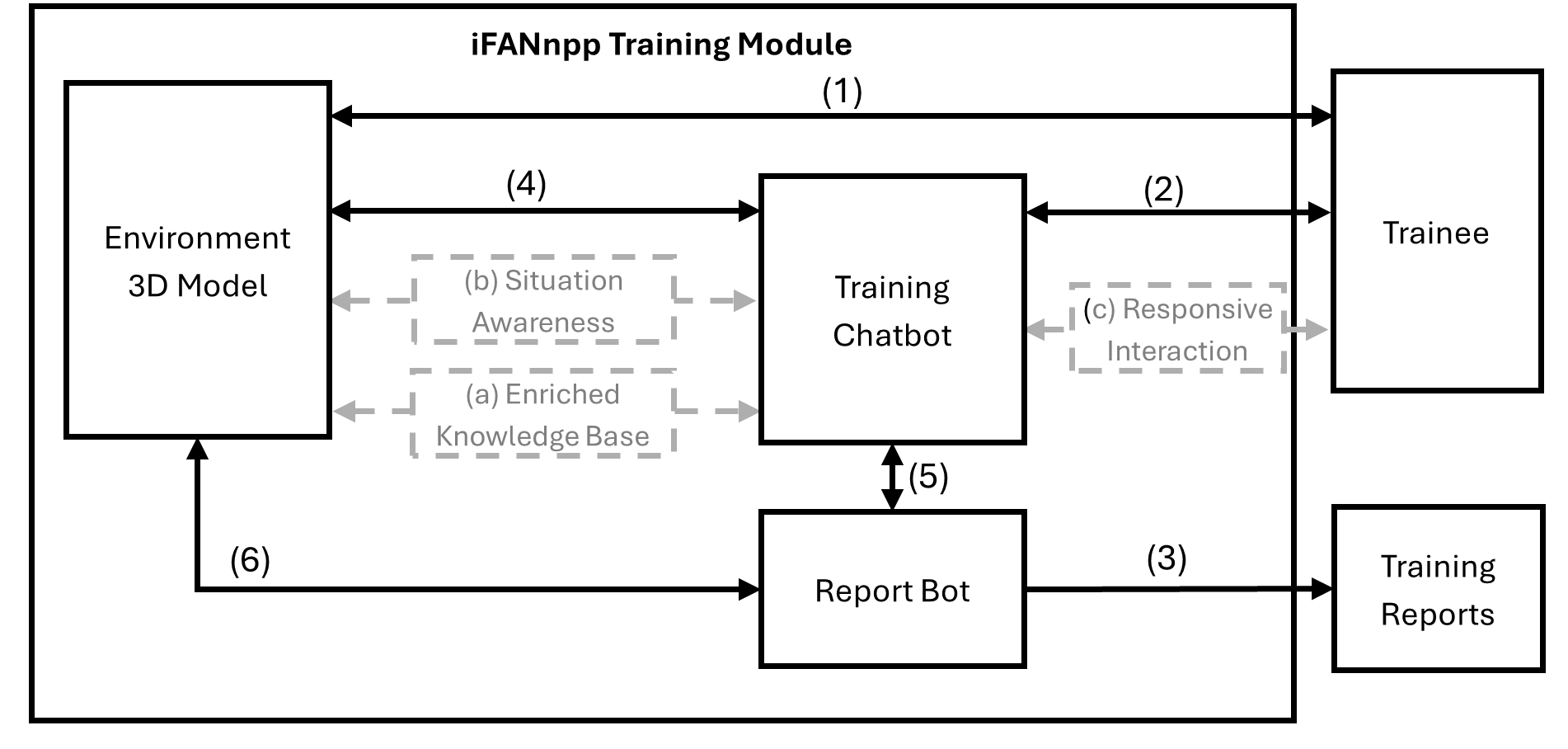}
    \caption{Structure of the Training Module}
    \label{fig:structure-training-module}
\end{figure*}

As illustrated in Figure \ref{fig:structure-training-module}, the functional modules related to workforce training consist of three primary components: the 3D environment, the training chatbot, and the reporting bot. The solid arcs with numbers in the figure indicate the major interaction between these components. In practice, trainees use a head-mounted display (HMD) to view the virtual environment and two hand controllers to interact with the virtual objects, as illustrated by arc (1). During the training session, the chatbot processes trainees’ actions and provides real-time guidance through a chat box in the trainees' view, as indicated by arc (2). Meanwhile, the training chatbot retrieves task-related information directly from the 3D environment, such as the trainee’s movement trajectory, the status of components being manipulated, and the operational states of other system elements, as shown in arc (4). In parallel, the report bot collects data from both the 3D environment shown in arc (6) and the training bot shown in arc (5) to generate the performance report. Once the training session concludes, the report bot finalizes the performance evaluation and provides the report to the trainees, as represented by arc (3).

Currently, more advanced training functions are under development, which include several key enhancements illustrated by the dashed boxes and arcs in Figure \ref{fig:structure-training-module}. First, an enriched knowledge base is being developed that covers the target system, training objectives, and additional contextual information. Second, the chat box is being enhanced with intelligent and situation-aware capabilities, allowing it to provide more constructive suggestions to trainees not only through text but also through speech.

\subsection{Teleoperation}
Teleoperation can reduce human exposure to radiation, heat, and spatially constrained environments while preserving the precision and situational awareness of an operator for NPP maintenance tasks. Beyond real operational jobs, teleoperation can also be used to train operators in a safe virtual environment before they directly interact with real systems, preventing negative consequences that may arise from operator errors or system failures. However, direct deployment of the teleoperation system into NPPs is limited, mainly due to safety concerns from non-optimized technology and unskilled operators without rigorous verification.

To address this limitation, the iFAN ecosystem serves as a middle ground between teleoperation development and real-world deployment by acting as the environment for engineers to exercise robot navigation and maintenance tasks under realistic settings without any physical risk. Within the SORT framework \cite{do2025teleoperation}, the iFAN ecosystem provides a place, where teleoperation control mechanisms, end-effector designs, and task procedures are tested and refined before transitioning to the physical system integration phase. This modularized approach allows hazardous cases, such as tight clearances in pipe clusters with high temperature gradients, to be explored safely and repeatedly. Also, as the iFAN ecosystem runs in real-time, latency, controller–robot alignment, and inverse kinematic behaviors can be evaluated under realistic motion and contact conditions. For example, robot behaviors such as overshoot, jitter, or poor camera placement can be identified and adjusted within the DT environment, reducing burdens, rework, and risks if the same teleoperation pipeline is deployed on hardware directly.

\begin{figure*}[t]
     \centering
        \includegraphics[width=\textwidth]{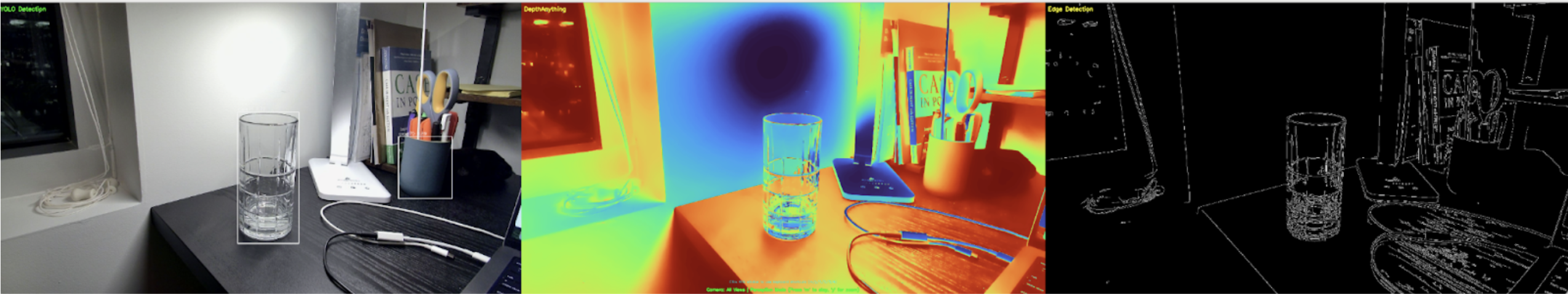}
    \caption{Example of a Multimodal Perception for VR}
    \label{fig:multimodalPerception}
\end{figure*}

Currently, the multimodality aspect in teleoperation is under development as it exposes operators to an enhanced environment and provides more task-relevant information than a single RGB camera can provide. For example, thermal camera perception data or edge detection projecting structural contours along with the primary camera view can help operators to understand which components are hot or where sharp geometric boundaries are located, as shown in Figure~\ref{fig:multimodalPerception}. By integrating multimodality into the current teleoperation framework, the iFAN ecosystem enables the simultaneous generation of semantic segmentation and diverse sensor channels, and it can serve as a more robust and comprehensive testbed.

\subsection{Robot Inspection}

To evaluate the iFAN ecosystem, two distinct robotic inspection scenarios are demonstrated: ground-based crack detection and aerial radiation monitoring. These demonstrations focus on the seamless integration of heterogeneous sensors and real-time vision systems. The ecosystem provides a safe environment for validating these technologies before they are introduced to real-world NPP settings directly. 

\begin{figure}[b]
    \centering
    \includegraphics[width=0.8\linewidth]{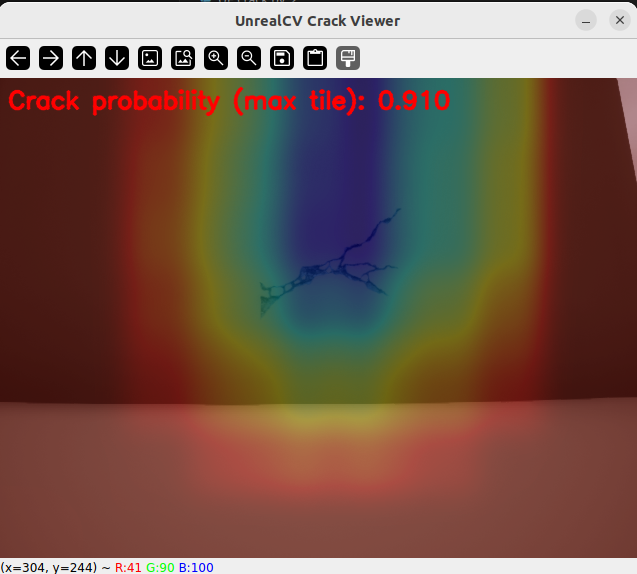}
    \caption{UnrealCV-Based Crack Detection}
    \label{fig:cracks}
\end{figure}

For the ground-based crack detection scenario, a wheeled robot is utilized. While the platform has built-in LiDAR, the iFAN ecosystem leverages the Rapyuta plugins \cite{RapyutarclUE, RapyutaSim} to provide custom sensors that stream information directly into ROS 2. The UE5 environment also allows for cameras to be added to any actor, giving TurtleBot 3 a camera for teleoperated crack inspection. As seen in Figure \ref{fig:cracks}, a crack detection ML model is used for inspection while the TurtleBot navigates the environment. 

This ML model is combined with the dosimeter and radiation source features to create a robot that roams the power plant to inspect systems with built-in radiation monitoring.

\begin{figure}[b]
    \centering
    \includegraphics[width=1\linewidth]{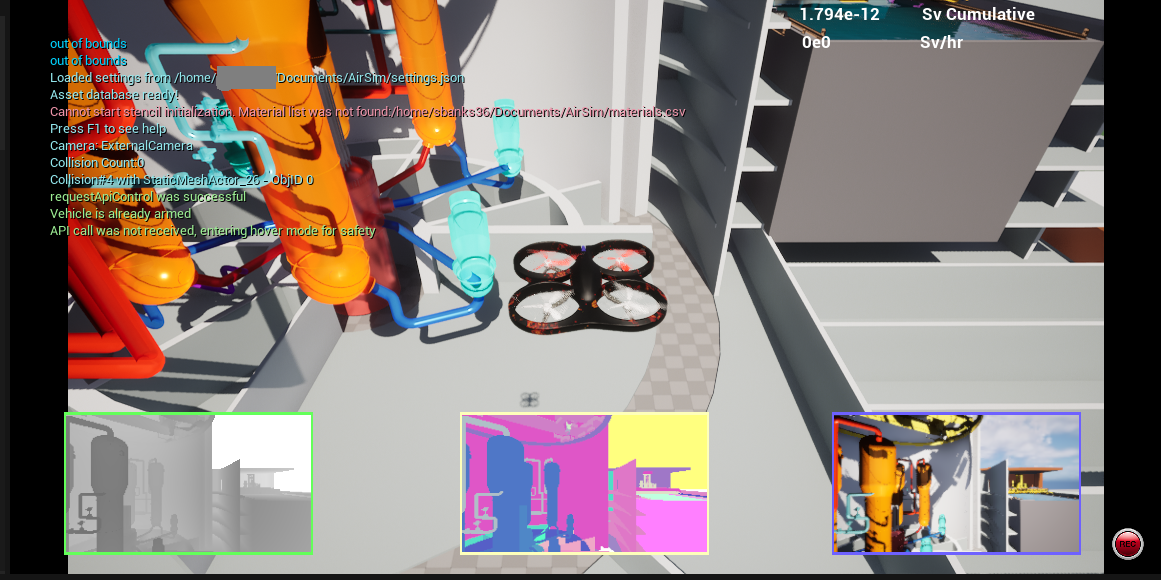}
    \caption{Cosys Airsim Drone Teleoperation}
    \label{fig:drone}
\end{figure}

The aerial inspection scenario utilizes the Airsim \cite{airsim2017fsr}, a Microsoft-developed physics simulator using UE4 as a training and testing environment. An updated version was used in this project, namely Cosys Airsim \cite{cosysairsim2023jansen}. This drone simulator is primarily for RL, with a gym environment connecting UE5 and Airsim directly. By default, the drone is a quadcoptor model, with a front-facing camera employing three modes: depth, segmentation, and RGB. The drone implemented in the iFAN ecosystem is also currently capable of teleoperation for a human pilot to utilize the drone for inspections, with cameras and a dosimeter to ensure safety as shown in Figure \ref{fig:drone}.

\subsection{Robot Navigation}
\label{sec:Robot Navigation}

Using robot models and RL features in the DT, robot path planning is demonstrated to validate its usage as a pre-deployment testbed. The task demonstrates the use of the iFAN ecosystem for testing a spatially-aware automated robot task, taking advantage of the unique geometries of the simulated power plant. It is an essential skill for mobile robots within a nuclear power plant to perform routine inspection and maintenance. 

\begin{figure}[t]
    \centering
    \includegraphics[width=1\linewidth]{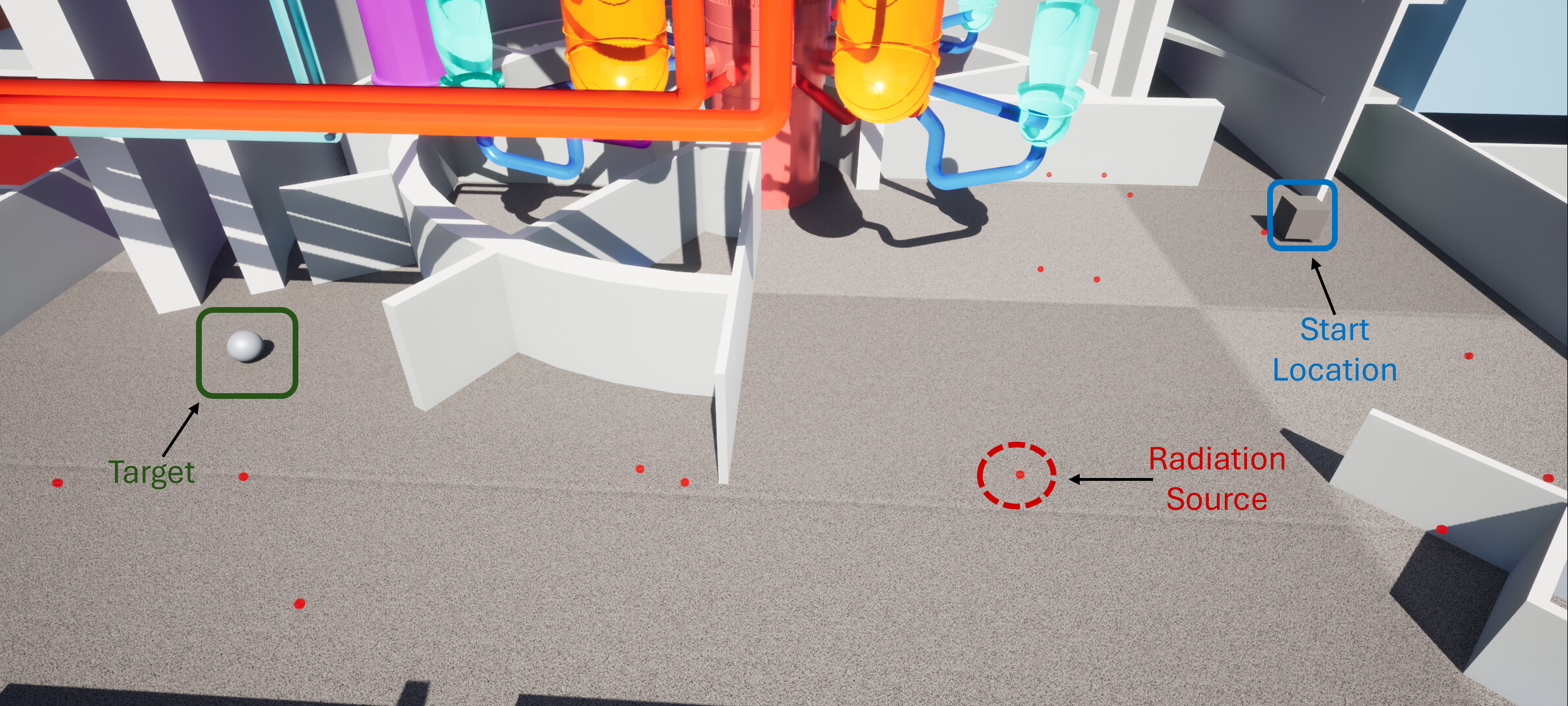}
    \caption{RL Navigation Project Setup}
    \label{fig:rl-setup}
\end{figure}

 Figure \ref{fig:rl-setup} visualizes the setup for the autonomous robot navigation task. The goal of this task is for the robot to navigate from a given start location to a target goal location while avoiding collision with the surrounding walls of the environment. In addition, the task requires the robot to avoid proximity to radiation, a challenge unique to nuclear environments containing ionizing radiation sources. The locations of the radiation sources are randomized every episode during both training and evaluation. Radiation sources are modeled, using the UE5 \textit{Debug Sphere} feature, and are visualized in red. The centers of the spheres represent radiation sources, with the diameters denoting areas deemed to be high radiation, to be avoided. The iFAN ecosystem is internally represented as a discretized grid-world environment, with each unit corresponding to 10 centimeters of distance within the digital twin. Therefore, locations are discrete from the perspective of the robot. Note that continuous position information can still be queried from the iFAN ecosystem environment, which can be used for alternate continuous robot control tasks.

To enable RL-based robot training, the key components of an RL environment are defined: a reward function to encourage positive decisions and discourage negative decisions, observations as defined input to the robot for decision making, and actions serving as the model output decisions at every time step.  The details of these components are as follows: 

\begin{itemize}
    \item \textbf{Reward} The reward is used to encourage the policy to reduce the Manhattan distance from the goal location while avoiding collisions with walls and radiation sources. The following reward function is used:
\[
    \text{Reward} = 
\begin{cases}
    10, & \text{If target reached} \\
   -0.1, & \text{If collided} \\
      0, & \text{If max steps reached} \\
    .99\Phi(s') - \Phi(s), & \text{Otherwise}
\end{cases}
\]

where $\Phi(s)$ is the negative Manhattan distance in 2D from the goal, normalized by the number of discrete locations $N$, defined in Equation 2:
\begin{equation}
    \Phi(s) = - \frac{|x_{goal} - x_{curr}| + |y_{goal} - y_{curr}|}{N}
\end{equation}

The difference function $.99\Phi(s') - \Phi(s)$ is a potential function \cite{ng1999policy}, used to prevent rewarding cyclical actions, such as iteratively moving towards and away from the goal location. 

\item \textbf{Observations}
The current location and relative target location is provided as observation, as well as a binary mask denoting the current surrounding area of the robot that is collidable. Additionally, locations where the robot has been since the restart of the episode are tracked. A binary mask denoting what parts of the surrounding area of the robot have been visited is also included as an observation. This provides the robot with historical information to help with decision-making, such as reducing repeated visits to the same location for a given run.

\item \textbf{Actions}
The action space is defined as a discrete set of movements within the 2D space. Specifically, the agent selects from four directions (up, down, left, or right)  and moves along the chosen direction. 

\end{itemize}

If the robot either collides with an obstacle or moves too close to a radiation source, the current episode is terminated, and the robot is reset to the starting position. The Double  Q-learning algorithm \cite{van2016deep} is used as the underlying RL algorithm for training. Epsilon-greedy exploration is used to produce a random action some proportion of the time during training to allow the robot to discover new areas of the map and the effects that actions have on them. This fraction is set to 50\% and decays to 1\% over time to encourage exploration earlier in the training phase. Upon running the example project training, it was observed that the final robot navigation model was able to reach the goal location from the start location with a 96\% success-rate, when evaluated deterministically across 50 episodes. This example gym environment serves as a template for creating future gym environments that support interaction with the iFAN ecosystem.

\subsection{Physical Security}
\label{sec:Physical Security}

Physical security is one of the most rigorously regulated system-critical domains in nuclear engineering. The target of physical security analysis is to prevent, detect, delay, and neutralize unauthorized access, malicious intrusions, and any sabotage actions that may result in catastrophic radiological consequences \cite{shubayr2024nuclear}. Traditional physical security analysis performed for physical protection systems (PPS) design heavily relies on static models, rule-based assessment frameworks, and field testing conducted under highly constrained conditions \cite{el2021analysis}. These methods lack the ability to capture the increasing complexity, dynamism, and heterogeneity of modern nuclear facilities. To address these technical and methodological gaps, the iFANnpp ecosystem offers several functionalities that integrate the 3D layout of an NPP with cyber-physical simulations, human-in-the-loop experiments, and robotics simulations.

\begin{figure}
    \centering
    \includegraphics[width=\linewidth]{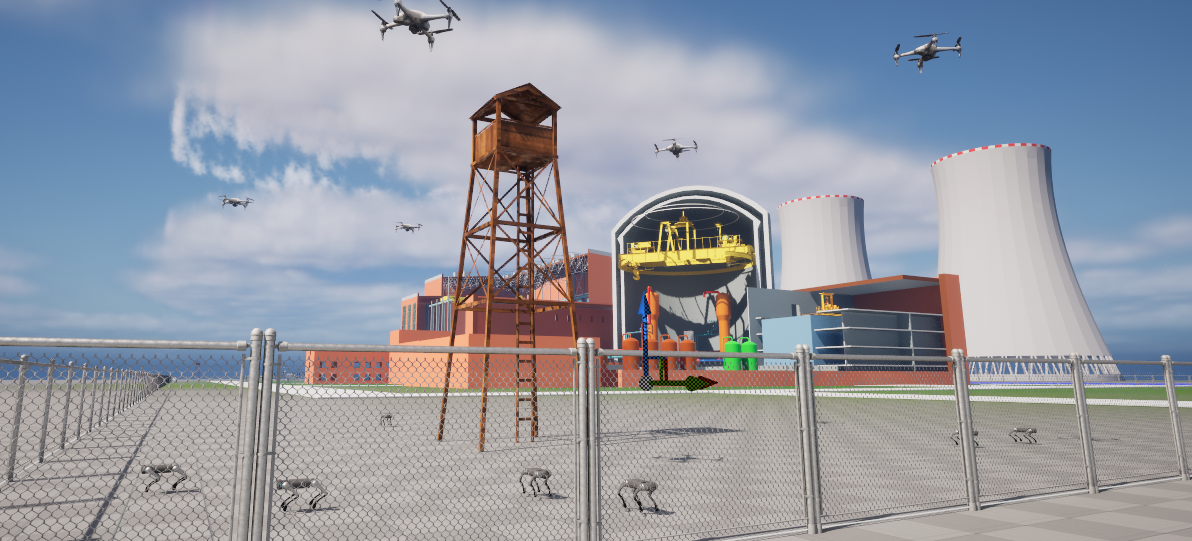}
    \caption{A Screenshot of Nuclear Facilities Modeled by the Ecosystem}
    \label{fig:screenshot-facilities}
\end{figure}

A high-fidelity 3D intrusion and defense simulation is currently under active development, including robots and drones, as shown in Figure \ref{fig:screenshot-facilities}. The 3D environment provided by the ecosystem realistically models an NPP’s spatial, structural, and functional layout, including internal plant architecture, such as the reactor containment, fuel-handling areas, the turbine building, and the switchyard. Additionally, the 3D model incorporates several perimeter barrier elements, including fences, vehicle exclusion zones, and guard towers. These essential security elements enable researchers to construct a multidimensional model of both the adversary pathways and the defensive response options. Beyond manual security oversight, the ecosystem facilitates the development, testing, and refinement of autonomous or semi-autonomous robotic agents for physical protection tasks across NPP sites. Through this development, intrusion simulations can be conducted that generate attacker and defender interception routes and calculate the performance delay barrier. These results are insightful for improving patrol routes, validating alarm-assessment workflows, or adjusting barrier configurations.

\subsection{Cybersecurity}
\label{sec:Demo-CyberSecurity}

A cyber-attack scenario was conducted to demonstrate how established data synchronization between a physical network and the iFAN DT can be used for scenario testing, enabling cybersecurity technique development for nuclear system hardening. For this scenario, a severe case was assumed, where an attacker has direct access to the control network of the circulating water system. An FDI attack was simulated on a circulating water system temperature sensor. The FDI serves as a representative industrial system attack referenced within the MITRE ATT\&CK framework for industrial control systems \cite{mitreattack}. To simulate this attack, the GPWR simulator was written a false value of 200 degrees Celsius. This information was then read into the physical testbed PLC, as described in section \ref{sec:Cybersecurity}, which subsequently synchronizes with the iFAN DT environment. 

\begin{figure}[t]
    \centering
    \includegraphics[width=1\linewidth]{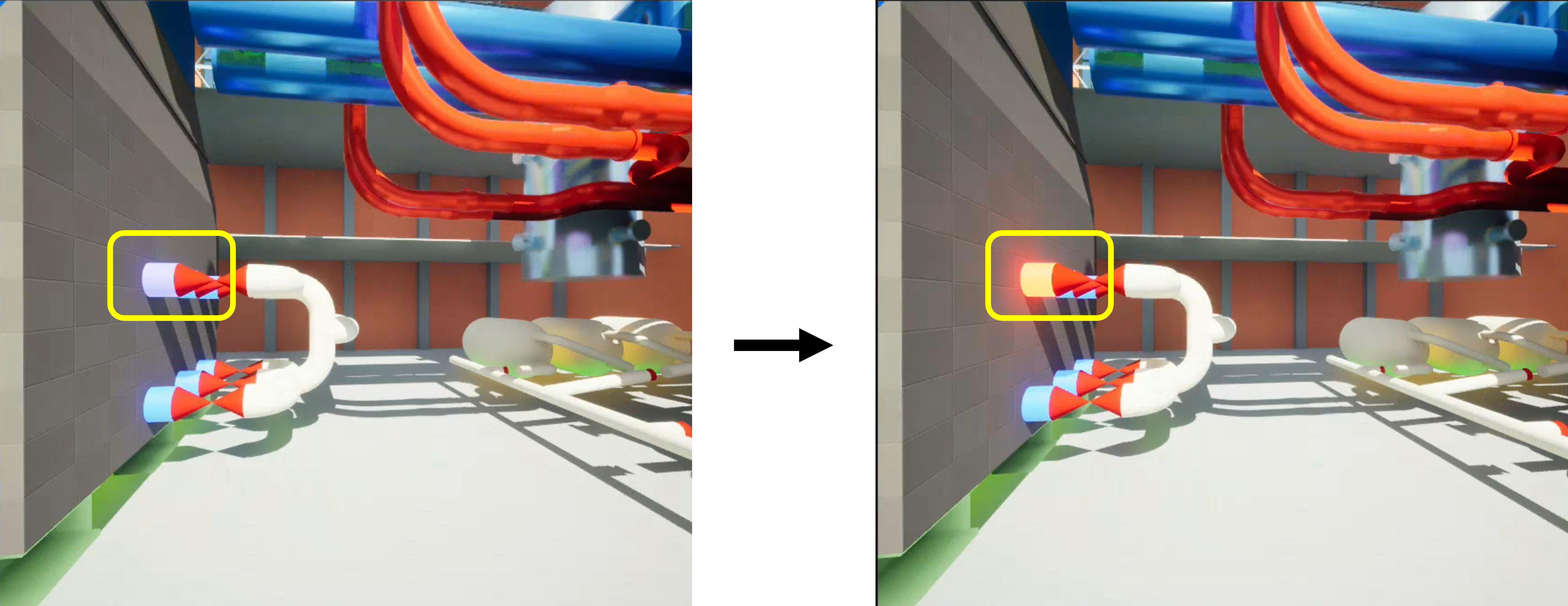}
    \caption{FDI Attack on Circulating Water System sensor. Typical value of $\sim 14.77$\textdegree C (left). FDI attack of 200\textdegree C (right)}
    \label{fig:testbed-comms-fdi}
\end{figure}

Figure \ref{fig:testbed-comms-fdi} displays visual results from the FDI attack scenario. Prior to the attack, the nominal temperature of $\sim 14.77$ \textdegree C is initially read into the iFAN DT through the testbed network from the PLC storing the sensor value read from GPWR. The temperature value is visually reflected in the DT with a purple color. Following the attack that modifies the relevant GPWR variable to 200\textdegree C, this information is propagated through the network, and a bright red is displayed around the sensor within the DT. Note that while the effect of the attack can be seen visually, the numerical value can additionally be explicitly queried within the DT environment for cyber-defense analysis. iFAN ecosystem's network DT thus provides necessary avenues to develop methods to assess, defend, and nullify threat axes in a controlled and safe environment to prevent possible disaster in a real-world scenario.

\subsection{Hardware-in-the-Loop}
One illustration of the HIL testbed's ability involves replacing the SG level control logic inside the GPWR simulator with an external function block program implemented in the PLC (Allen-Bradley ControlLogix). Once the simulator shares SG-related process variables and status signals over the FactoryTalk Linx OPC server, they are read into the controller scope tags in the PLC over Common Industrial Protocol (CIP). The controller then executes the three element SG level control functions for feedwater flow, steam flow, and SG level, finally sending the resulting actuator commands back through the OPC to drive the feedwater valves and pumps.

PLC synchronization logic is implemented for timing coordination, where GPWR runs with a 50 ms integration step and after each step pushes updated process variables and overall simulator status to the PLC. A dedicated synchronization routine in the PLC is executed in the same 100 Hz periodic task as the SG control logic, using the status tags to enable, freeze, or reset instructions and internal states such that the PLC does not integrate or perform other computations when GPWR is paused or reset. This could otherwise lead to runaway control outputs and unrealistic transients if not addressed. The PLC scan rate is intentionally set faster than the simulator step rate to minimize missed updates while accepting small, explainable discrepancies due to different numerical precision and filtering implementations.

Previous work \cite{chen2024full} on the HIL testbed involved embedding malicious routines in the PLC logic that can manipulate sensor inputs or actuator commands to induce low-amplitude oscillations. This type of behavior does not directly trip nuclear safety-related protections, but it does cause distinctive dynamics in selected process variables which would be undesirable or challenging to operations. In the HIL environment, these attacks remain largely invisible in key indicators such as reactor power and SG level but produce damaging high-frequency oscillations in components like the feedwater pumps.

This example illustrates how stealthy supply chain attacks can degrade equipment without obvious operator alarms. Because the testbed logs both plant process data and network traffic, it can also be extended to evaluate detection algorithms, assess resilience under scenarios such as denial-of-service, replay attacks, and false data injections. Additionally, it can help differentiate between HIL architectures in terms of realism, scalability, and benefit to research efforts.

\section{Conclusion}
\label{sec:Conclusion}

The nuclear industry is undergoing a paradigm shift, transitioning to a digital instrumentation and control systems, and incorporating emerging technologies such as remote operations and (near) autonomous operations. As demonstrated throughout this paper, the iFAN ecosystem serves as an essential test and evaluation platform for these new technologies, as it addresses both robotic operations and digital security. To support these two main fields, the ecosystem provides a safe pipeline with several core features designed for nuclear environments.

The core features enable a wide range of specialized research and operational applications. Virtual reality capabilities provide both operator training and robotic teleoperation, enhancing knowledge and safe execution of operations in NPPs. Reinforcement learning models upgrade robotic capabilities for their deployment in a power plant. Artificial intelligence and machine learning integration are used to find new ways to optimize plant and system performance. The expanding use of advanced digital system necessitates a strong employment of cybersecurity techniques. Having a cyber-physical setup ensures effective testing and development of new defenses against possible cyber-threats. Radiation simulation has boundless utility for everyday and emergency operations training and planning. Hardware-in-the-loop research works in tandem with monitoring of plant performance following cyber-attacks, leading to ever more increased defense capabilities. The iFAN ecosystem has boundless opportunities to investigate and design some of the newest technologies offered today, and with its evolving design, it has the versatility to take on even more research to assist in modernizing the nuclear field. 

\section*{Acknowledgment}
This material is based upon work supported by the U.S. Department of Energy, Office of Nuclear Energy, Distinguished Early Career Award, under Award number DE-NE0009306 , and Nuclear Energy University Program (NEUP), under Award Number DE-NE0009443.





\end{document}